\documentclass{llncs}

\usepackage{graphicx}
\usepackage{amsmath,amsfonts}

\usepackage{listings}
\newcommand{\sysfont}{\textit}
\newcommand{\clingo}{\sysfont{clingo}}
\newcommand{\ie}{\emph{i.e.},~}

\newcommand{\eg}{\emph{e.g.}~}
\newcommand{\al}{\emph{al.}~}
\renewcommand{\vec}[1]{\ensuremath{\boldsymbol{#1}}}

\begin{document}

\title{Declarative sequential pattern \\mining of care pathways\thanks{This research is supported by the PEPS project funded by the french agency for health products safety (ANSM) and the SePaDec project funded by Brittany Region.}}

\author{Thomas Guyet\inst{1} \and Andr\'e Happe\inst{2} \and Yann Dauxais\inst{3}}
\authorrunning{Thomas Guyet et al.}

\institute{AGROCAMPUS-OUEST/IRISA-UMR6074
\and
CHRU Brest/EA-7449 REPERES
\and
Rennes University 1/IRISA-UMR6074
}

\maketitle

\begin{abstract}
Sequential pattern mining algorithms are widely used to explore care pathways database, but they generate a deluge of patterns, mostly redundant or useless. Clinicians need tools to express complex mining queries in order to generate less but more significant patterns.
These algorithms are not versatile enough to answer complex clinician queries. 
This article proposes to apply a declarative pattern mining approach based on Answer Set Programming paradigm. It is exemplified by a pharmaco-epidemiological study investigating the possible association between hospitalization for seizure and antiepileptic drug switch from a french medico-administrative database.
\keywords{answer set programming, epidemiology, medico-administra\-tive databases, patient care pathways}
\end{abstract}

\section{Introduction}
Pharmaco-epidemiology applies the methodologies developed in general epidemiology 
to answer questions about the uses of health products in the population in real condition. 
In pharmaco-epidemiology studies, people who share common characteristics are recruited. Then, a dataset is built from meaningful data (drug exposures, events or outcomes) collected within a defined period of time. Finally, a statistical analysis highlights the links (or the lack of link) between drug exposures and outcomes (\eg adverse effects).

The data collection of such prospective cohort studies is slow and cumbersome. Medico-administrative databases are readily available and cover a large population. They record, with some level of details, all reimbursed drug deliveries and all medical procedures, for insured people. Such database gives an abstract view on longitudinal care pathways. It has become a credible alternative for pharmaco-epidemiological studies \cite{martin2010pharmacoepidemiological}. However, it has been conceived for administrative purposes and their use in epidemiology is complex. 

Our objective is to propose a versatile pattern mining approach that extracts sequential patterns from care pathways.  
The flexibility of such new knowledge discovery tools has to enable epidemiologists to easily investigate various types of interesting patterns, \eg frequent, rare, closed or emerging patterns, and possibly new ones. 
On the other hand, the definition of interesting patterns has to exploit in-depth the semantic richness of care pathways due to complex care event descriptions (\eg units number, strength per unit, drugs and diagnosis taxonomies, etc.).
By this mean, we expect to extract less but more significant patterns.

This article presents the application of a declarative pattern mining framework based on Answer Set Programming (ASP) \cite{Guyet_IJCAI2016} to achieve care pathway analysis answering pharmaco-epidemiological questions.

Answer Set Programming (ASP) is a declarative programming paradigm. It gives a description, in a first-order logic syntax, of what is a problem instead of specifying how to solve it.
Semantically, an ASP program induces a collection of so-called \emph{answer sets}.
For short, a model assigns a truth value to each propositional atoms of the program. An answer set is a minimal set of true propositional atoms that satisfies all the program rules.
ASP problem solving is ensured by efficient solvers. For its computational efficiency, we use \clingo~\cite{gekakaosscsc11a} as a primary tool for designing our encodings. 
An \emph{ASP program} is a set of rules of the form:  $\tt a_0 \mathrel{\texttt{:-}} a_1,\dots,a_m, \text{\lstinline!not! } a_{m+1},\dots,\text{\lstinline!not! } a_n$, 
where each $\mathtt{a_i}$ is a propositional atom for $\mathtt{0}\leq\mathtt{i}\leq\mathtt{n}$ and \lstinline!not! stands for \emph{default negation}. In the body of the rule, commas denote conjunctions between atoms.
If $\mathtt{n}=0$, \ie the rule body is empty, the rule is called a \emph{fact} and the symbol ``\texttt{:-}'' may be omitted. Such a rule states that the atom $a_0$ has to be true.
If $\mathtt{a_0}$ is omitted, \ie the rule head is empty, the rule represents an integrity constraint meaning that it must not be \emph{true}.
\clingo\ also includes several extensions to facilitate the practical use of ASP (variables, conditional literals and cardinality constraints).

Recent researches has been focused on the use of declarative paradigms, including ASP, to mine structured datasets, and more especially sequences \cite{Guyet_IJCAI2016,Negrevergne15}. The principle of declarative pattern mining is closely related to the Inductive Logic Programming (ILP) \cite{quiniou01} approach. The principle is to use a declarative language to model the analysis task: supervised learning for ILP and pattern mining for our framework. The encoding benefits from the versatility of declarative approaches and offers 
natural abilities to represent and reason about knowledge.

\section{Context, data and pharmaco-epidemiological question}\label{sec:data}
In this work, we exemplify our declarative pattern mining framework by investigating the possible association between hospitalization for seizure and antiepileptic drug switches, \ie changes between drugs.
The first step was to create a digital cohort of 8,379 patients with a stable treatment for epilepsy (stability criterion detailed in \cite{polard2015brand} have been used).
This cohort has been built from the medico-administrative database, called SNIIRAM \cite{martin2010pharmacoepidemiological} which is the database
 of the french health insurance system. It is made of all outpatient reimbursed health expenditures. Our dataset represents 1,8M deliveries of 7,693 different drugs and 20,686 seizure-related hospitalizations.

This dataset and background knowledge (ATC drugs taxonomy, ICD-10 taxonomy) are encoded as ASP facts.
For each patient $\vec{p}$, drug deliveries are encoded with \lstinline!deliv($\vec{p}$,$t$,$d$,$q$)! atoms meaning that patient $\vec{p}$ got $q$ deliveries of drug $d$ at date $t$. Dates are day numbers starting from the first event date. We use french CIP, Presentation Identifying Code, as drug identifier. The knowledge base links the CIP to the ATC and other related informations (\eg speciality group, strength per unit, number of units or volume, generic/brand-named status, etc).
Each diagnosis related to an hospital stay is encoded with \lstinline!disease($\vec{p}$,$t$,$d$)! meaning that patient $\vec{p}$ have been diagnosed with $d$ at date $t$.
Data, $\mathcal{D}$, and related knowledge base, $\mathcal{K}$, represent a total of 2,010,556 facts.

\section{Sequential pattern mining with ASP}
Let $\mathcal{I}=\{i_1,i_2,\ldots,i_{|\mathcal{I}|}\}$ be a set of \emph{items}.
A \emph{temporal sequence} $\vec{s}$, denoted by $\left\langle \left(s_j, t_j\right) \right\rangle_{j \in [m]}$ is an ordered list of items $s_j \in \mathcal I$ timestamped with $t_j \in \mathbb{N}$. 
Let $\vec{p}=\langle p_j\rangle_{1\leq j\leq n}$, where $ p_j \in \mathcal{I}$ be a sequential pattern. We denote by $\mathcal{L}=\mathcal{I}^*$ the \emph{pattern search space}.
Given the pattern $\vec{p}$ and the sequence $\vec{s}$ with $n \le m$,  we say that $\vec{s}$ \emph{supports} $\vec{p}$, iff there exists $n$ integers $e_1<\ldots<e_n$ such that $p_k = s_{e_k},\; \forall k\in \{1,\ldots,n\}$. $(e_k)_{k\in[n]}$ is called an \emph{embedding} of pattern $\vec{p}$ in $\vec{s}$. $\mathcal{E_{\vec{p}}^{\vec{s}}}=\{(e_k)_{k\in[n]}\}$ denotes the set of the embeddings of $\vec{p}$ in $\vec{s}$.
Let $\mathcal{D}=\left\{ \vec{s}^k \right\}_{k \in [N]}$, be a dataset of $N$ sequences. We denote by $\mathcal{T}_{\vec{p}}$ the sequence set supported by $\vec{p}$.
Given a set of constraints $\mathcal{C}$, the mining of sequential patterns consists in finding out all tuples $\langle\vec{p},\mathcal{T}_{\vec{p}},\mathcal{E}_{\vec{p}}\rangle$ satisfying $\mathcal{C}$, where $\mathcal{E}_{\vec{p}}=\bigcup_{s\in\mathcal{T}_{\vec{p}}} \mathcal{E}_{\vec{p}}^{\vec{s}}$. The most used pattern constraint is the minimal frequency constraint, $c_{f_{min}}:\left|\mathcal{T}_{\vec{p}}\right|\geq f_{min}$, saying that the pattern support has to be above a given threshold $f_{min}$.

Sequential pattern mining with ASP has been introduced by Guyet et \al \cite{Guyet_IJCAI2016}\footnote{Original encodings can be found here: \url{https://sites.google.com/site/aspseqmining/}}. It encodes the sequential pattern mining task as an ASP program that process sequential data encoded as ASP facts.
A sequential pattern mining task is a tuple $\langle \mathcal{S}, \mathfrak{M}, \mathcal{C} \rangle$, where $\mathcal{S}$ is a set of ASP facts encodings the sequence database, $\mathfrak{M}$ is a set of ASP rules which yields pattern tuples from database, $\mathcal{C}$ is a set of constraints (see \cite{Negrevergne15} for constraint taxonomy). We have $\mathcal{S} \cup \mathfrak{M} \cup \mathcal{C} \models \left\{\left\langle\vec{p},\mathcal{T}_{\vec{p}},\mathcal{E}_{\vec{p}}\right\rangle\right\}$.

In our framework, the sequence database is modeled by \lstinline!seq($\vec{s}$,$t$,$e$)! atoms. Each of these atoms specifies that the event $e\in\mathcal{I}$ occurred at time $t$ in sequence $\vec{s}$. 
On the other hand, each answer set holds atoms that encode a pattern tuples. \lstinline!pat($i$,$p_i$)! atoms encode the pattern $\vec{p}=\langle p_i \rangle_i\in[l]$ where $l$ is given by \lstinline!patlen($l$)!,  \lstinline!support($\vec{s}$)! encodes $\mathcal{T}_{\vec{s}}$ 
and finally $\mathcal{E}_{\vec{p}}$ is encoded by \lstinline!occ($\vec{s}$,$i$,$e_i$)! atoms.

\section{Declarative care pathway mining}\label{sec:principle}
The declarative care pathway mining task can be defined as a tuple of ASP rule sets $\langle \mathcal{D}, \mathfrak{S}, \mathcal{K}, \mathfrak{M}, \mathcal{C} \rangle$ where $\mathcal{D}$ is the raw dataset and $\mathcal{K}$ the knowledge base introduced in section \ref{sec:data}; $\mathfrak{M}$ is the encoding of the sequence mining task presented in \cite{Guyet_IJCAI2016} and $\mathcal{C}$ is a set of constraints. Finally, $\mathfrak{S}$ is a set of rules yielding the sequences database: $\mathfrak{S} \cup \mathcal{D} \cup \mathcal{K} \models \mathcal{S}$.
Depending on the study, the expert has to provide $\mathfrak{S}$, a set of rules that specifies which are the events of interest and $\mathcal{C}$, a set of constraints that specifies the patterns the user would like.

In the following of this section, we give examples for $\mathfrak{S}$ and $\mathcal{C}$ to design a new mining tasks inspired from a \emph{case-crossover study} answering our clinical question~\cite{polard2015brand}. For each patient, the $\mathfrak{S}$ rules generate two sequences made of deliveries within respectively the 3 months before the first seizure (positive sequence) and the 3 to 6 months before the first seizure (negative sequence). In this setting the patient serves as its own control.
The mining query consists in extracting frequent sequential patterns where a patient is supported by the pattern iff the pattern appears in its positive sequence, but not in its negative sequence. 
A frequency threshold for this pattern is set up to 20 and we also constraint patterns 1) to have generic and brand-name deliveries and 2) to have exactly one switch from a generic to a brand-name anti-epileptic grugs -- AED (or the reverse).

\paragraph{Defining sequences to mine with $\mathfrak{S}$.}
Listing above illustrates the sequence generation of deliveries of anti-epileptic drug specialities within the 3 months (90 days) before the first seizure event. It illustrates the use of the knowledge base to express complex sequences generation.
In this listing, \lstinline!aed($i$,$c$)! lists the CIP code $i$, which are related to one of the ATC codes for AED ($N03AX09$, $N03AX14$, etc.), and \lstinline!firstseizure($\vec{p}$,$t$)! is the date, $t$, of the first seizure of patient $\vec{p}$. A seizure event is a disease event with one of the G40-G41 ICD-10 code. The first seizure is the one without any other seizure event before.
ASP enables to use a reified model of sequence where events are functional literals. \lstinline!seq(P,T,deliv(AED,Gr,G))! designates that patient \lstinline!P! was delivered at time \lstinline!T! with a drug where \lstinline!AED! is the ATC code, \lstinline!Gr! identify the drug speciality and \lstinline!G! indicates whether the speciality is a generic drug or a brand-named one.
The same encoding can be adapted for sequences within the 3 to 6 months before the first seizure event.
\begin{lstlisting}[numbers=none,basicstyle=\ttfamily\scriptsize]
aed(CIP,AED):-cip_atc(CIP,AED),AED=(n03ax09;n03ax14;n03ax11;n03ag01;n03af01).
firstseizure(P,T) :- disease(P,T,D), is_a(D,g40;g41), 
                     #count{Tp: disease(P,Tp,Dp), is_a(Dp,g40;g41), Tp<T}=0.

seq(P,T,deliv(AED,Gr,1)):- deliv(P,T,CIP,Q), aed(CIP,AED), grs(CIP,Gr),
                             generic(CIP), T<Ts, T>Ts-90, firstseizure(P,Ts).

seq(P,T,deliv(AED,Gr,0)):- deliv(P,T,CIP,Q), aed(CIP,AED), grs(CIP,Gr),
                         not generic(CIP), T<Ts, T>Ts-90, firstseizure(P,Ts).
\end{lstlisting}

\paragraph{Defining constraints on patterns.} On the other side of our framework, $\mathcal{C}$ enables to add constraints on patterns the clinician looks for. Lines 1-3 (see listing above) encode the case-crossover constraints. They select patterns (\ie answer sets) that are frequently in the 3 months period but not in the 3 to 6 months period. The frequency threshold is set to 20.
Finally, lines 5-6 illustrate a constraint on the shape of the pattern, that here must contains exactly one switch from a brand-name to a generic drug (or the reverse).
\begin{lstlisting}[basicstyle=\ttfamily\scriptsize]
discr(T) :- support(T), not neg_support(T).
#const th=20.
:- { discr(T) } < th.
 
change(X) :- pat(X+1,deliv(AEDp,GRSp,Gp)), pat(X,deliv(AED,GRS,G)), Gp!=G.
:- #count{X:change(X)}!=1.
\end{lstlisting}

\paragraph{Results}
The solver extracts respectively 32 patterns and 21 patterns (against 4359 patterns with a regular sequential pattern mining algorithm). With such very constrained problem, the solver is very efficient and extracts all patterns in less than 30 seconds. 
The following pattern is representative of our results: {\small $\langle (N03AG01,438,1), (N03AG01,438,1), (N03AX14,1023,0), (N03AX14,1023,0) \rangle$} is a sequence of deliveries showing a change of treatment from a generic drug of the speciality $438$ of valproic acid to the brand-name speciality $1023$ of levetiracetam.
According to our mining query, we found more than 20 patients which have this care sequence within the 3 months before a seizure, but not in the 3 previous months preceding this period. These new hypothesis of care-sequences are good candidates for further investigations and possible recommendation about AE treatments.

\section{Conclusion}
Declarative sequential pattern mining with ASP is an interesting framework to flexibly design care-pathway mining queries that supports knowledge reasoning (taxonomy and temporal reasoning). 
We illustrated the expressive power of this framework by designing a new mining tasks inspired from case-crossover studies and shown its utility for care pathway analytics.
We strongly believe that our integrated and flexible framework empowers the clinician to quickly evaluate various pattern constraints and that it limits tedious pre-processing phases. 

\bibliographystyle{splncs}

\end{document}